 \documentclass[conference, 10pt]{IEEEtran}

\usepackage{times}
\usepackage[numbers]{natbib}
\usepackage{amsthm}
\usepackage{multicol}
\usepackage[bookmarks=true]{hyperref}
\usepackage{amsmath}
\usepackage[dvipsnames]{xcolor}
\usepackage{graphicx}
\usepackage[footnotesize]{caption}
\usepackage{subcaption}
\usepackage{amssymb}
\usepackage{array}
\usepackage{url}
\usepackage{color}
\usepackage{nccmath}
\usepackage[normalem]{ulem}
\usepackage[linesnumbered,ruled,vlined]{algorithm2e}\usepackage{enumerate}
\usepackage{multirow}
\usepackage{bbm}
\usepackage{marginnote}
\usepackage[T1]{fontenc}
\usepackage{mathtools}
\usepackage[letterpaper, left=0.667in, right=0.667in, bottom=0.625in, top=0.792in]{geometry}
 \usepackage{nopageno}
\usepackage{algorithmic}

\IEEEoverridecommandlockouts
\renewcommand{\theenumi}{\roman{enumi}}%

\SetCommentSty{mycommfont}

\SetKwInput{KwInput}{Initialize}
\SetKwInput{KwOutput}{Output}
\SetKwInOut{Requires}{Requires}
\SetKwInOut{Return}{Return}
\SetKwInOut{Algorithm}{Algorithm}

\DeclareMathOperator{\EX}{\mathbb{E}}

\newtheorem{defn}{Definition}
\newtheorem{rem}[defn]{Remark}
\newtheorem{lem}[defn]{Lemma}
\newtheorem{prop}[defn]{Proposition}
\newtheorem{assum}[defn]{Assumption}

\newtheorem{thm}[defn]{Theorem}
\newtheorem{cor}[defn]{Corollary}

\newcommand{\R}{\mathbb{R}}
\newcommand{\N}{\mathcal{N}}
\newcommand{\NN}{\mathbb{N}}
\newcommand{\Z}{\mathcal{Z}}
\newcommand{\W}{\mathcal{W}}
\newcommand{\I}{\mathcal{I}}

\newcommand{\J}{\mathcal{J}}
\newcommand{\K}{\mathcal{K}}
\newcommand{\E}{\mathcal{E}}
\newcommand{\X}{\mathcal{X}}

\newcommand{\Gauss}{\mathcal{N}}
\newcommand{\Oego}{\mathcal{O}^{ego}}
\newcommand{\Oobs}{\mathcal{O}^{obs}_i}
\newcommand{\Oobsset}{\mathcal{O}^{obs}}

\providecommand{\Int}{\texttt{int}}

\newcommand{\zonocg}[2]{<{#1},{#2}>}
\providecommand{\tz}{t_0}
\providecommand{\tplan}{t_\text{plan}}
\providecommand{\tnb}{t_\text{m}}
\providecommand{\tf}{t_{f}}

\providecommand{\opt}{(\texttt{Opt}) }
\providecommand{\optE}{(\texttt{Opt-E}) }
\providecommand{\rot}{\texttt{rotate}}
\providecommand{\kstar}{k^{*}}

\newcommand{\JW}[1]{\textnormal{{\textbf{}}}}
\newcommand{\Ram}[1]{\textnormal{{\textbf{}}}}

\pagebreak

\begin{document}

\title{Not All Actions Are Created Equal: Bayesian Optimal Experimental Design for Safe and Optimal Nonlinear System Identification}

\author{Parker Ewen$^{1}$, Gitesh Gunjal$^{1}$, Joey Wilson$^{1}$,  Jinsun Liu$^{1}$, Challen Enninful Adu$^{1}$, and Ram Vasudevan$^{1}$
\thanks{All authors affiliated with the Robotics Department or the Mechanical Engineering Department at the University of Michigan, 2505 Hayward Street, Ann Arbor, Michigan, USA, \{\tt \small pewen, gitesh, wilsoniv, jinsunl, enninful, ramv\} {\tt \small @umich.edu}}%
}

\maketitle

\begin{abstract}
Uncertainty in state or model parameters is common in robotics and typically handled by acquiring system measurements that yield information about the uncertain quantities of interest.
Inputs to a nonlinear dynamical system yield outcomes that produce varying amounts of information about the underlying uncertain parameters of the system.
To maximize information gained with respect to these uncertain parameters we present a Bayesian approach to data collection for system identification called Bayesian Optimal Experimental Design (BOED).
The formulation uses parameterized trajectories and cubature to compute maximally informative system trajectories which obtain as much information as possible about unknown system parameters while also ensuring safety under mild assumptions.
The proposed method is applicable to non-linear and non-Gaussian systems and is applied to a high-fidelity vehicle model from the literature.
It is shown the proposed approach requires orders of magnitude fewer samples compared to state-of-the-art BOED algorithms from the literature while simultaneously providing safety guarantees.
\end{abstract}

\section{Introduction} \label{sec:intro}
Robots operating in the world must contend with noisy sensor measurements, uncertainty in their environment representations, and intrinsic uncertainty in their dynamics model.
Although noise may be minimized with more precise sensors or improved algorithms, uncertainty persists and cannot be entirely eliminated.
Therefore, robots must be able to adapt to noise in their world model and intrinsic parameters while still operating in a safe manner.

The challenge of uncertainty is commonplace in robotics and is typically handled through Bayesian inference which describes how to update uncertain variables recursively using measurements from the dynamical system as it evolves.
State estimation methods such as the Kalman Filter and its variants update state estimates in closed-form assuming a Gaussian prior  \cite{barrau2016invariant, wan2001unscented}.
In the case of non-parametric state estimation, one can apply a sampling-based alternative such as the Particle Filter and its variants \cite{van2000unscented, carpenter1999improved}.

Dynamical systems may also contain uncertainty with respect to parameters in their dynamics, and must contend with this uncertainty alongside state uncertainty.
For instance, an autonomous vehicle may update its state estimate as well as its estimate for the coefficient of friction of the road.
The so-called \textit{dual} versions of Kalman Filter variants \cite{popovici2017dual, wan2001dual} and Particle Filter variants \cite{azam2012dual} are used to accomplish this task.

\begin{figure} [!t]
    \centering
.    \includegraphics[width=\columnwidth]{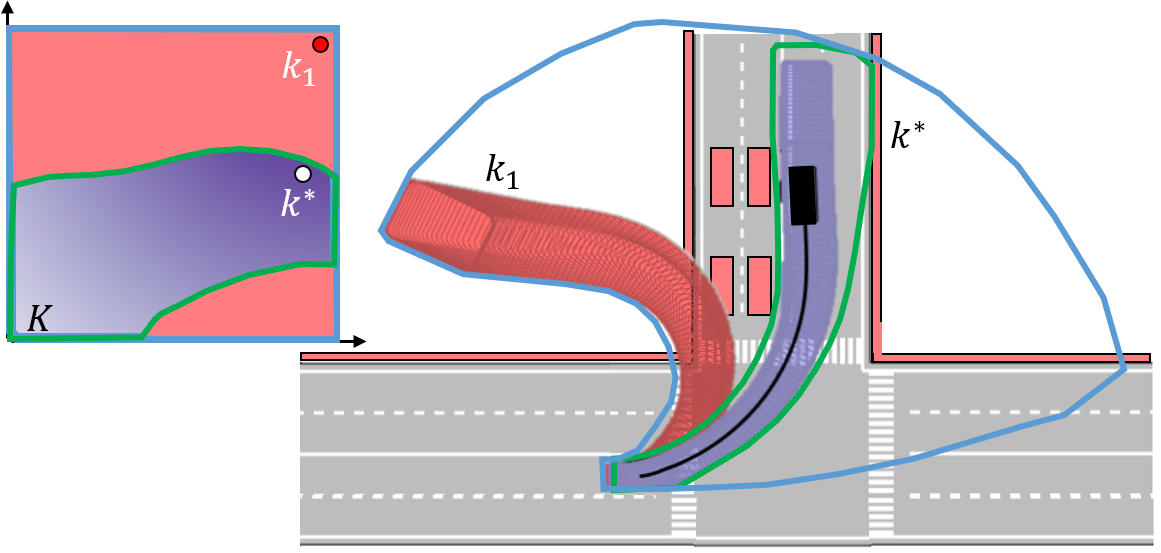}
    \caption{A demonstration of the safe, maximally informative system trajectory using the Bayesian Optimal Experimental Design approach described in this paper.
   The proposed method for computing optimally-informative trajectories relies on a novel differentiable approximation to the expected information gain for a dynamical system and includes safety guarantees, under mild assumptions, using reachability analysis.
    The proposed algorithm is demonstrated to compute system trajectories that minimize the entropy of the posterior system parameter estimates after one planning iteration.
    The blue polytope represents the union of all forward-reachable sets while the green polytope represents the subset of these that are not in collision with obstacles.
    The optimal trajectory parameter, $k^{*}$, which accounts for safety and its corresponding forward reachable set is shown in purple.
    An example of the optimally-informative trajectory without safety guarantees is depicted in red and parameterized using $k_1$.}
    \label{fig:main_figure}
\end{figure}

Often, inference algorithms in robotics take a passive approach to updating parameter estimates.
In reality, the actions a robot takes impact the amount of information obtained about unknown model parameters.
The study of designing experiments to learn information as efficiently as possible is known as Bayesian Optimal Experimental Design, and has seldom been explored for robotic system identification \cite{cooper2021bayesian}.

Efficiently learning unknown parameters in a timely and safe manner can have a significant impact on robots operating in an uncertain environment.
This paper explores how system inputs affect the posterior estimates of model parameters for nonlinear systems and presents an optimal planning algorithm that minimizes entropy of subsequent model parameter estimates.
Unfortunately selecting arbitrary maximally informative actions may sacrifice the overall safety of the robot system.
As a result, we also illustrate how to formulate a single trajectory optimization problem that identifies safe, maximally informative actions as illustrated in Figure \ref{fig:main_figure}. 
To ensure safety, we employ reachability analysis.

We demonstrate our approach on a non-linear model from the literature and on a high-fidelity autonomous vehicle model.
In both instances, we compare our approach for a single planning iteration against state-of-the-art techniques for Bayesian Optimal Experimental Design and illustrate the advantages of our differentiable and safe approach over other tested methods.
The contributions of this work are as follows:
\begin{itemize}
    \item A computationally tractable, differentiable approximation for the expected information gain metric used in experimental design; and
    \item A planning algorithm to compute a safe, optimally-informative trajectory for robot system identification.
\end{itemize}

The remainder of this paper is organized as follows:
Section \ref{sec:related} discusses related work in system safety under uncertainty and Bayesian Optimal Experimental Design;
Section \ref{sec:prelim} reviews preliminary concepts used throughout the paper;
Section \ref{sec:model} defines the dynamic and measurement model of the robot;
Section \ref{sec:bed} introduces definitions of information gain as well as the expected information gain metric used in Bayesian Optimal Experimental Design;
Section \ref{sec:alg} formulates our tractable, differentiable approximation of expected information gain and demonstrates how we incorporate safety into the planning framework;
Finally, Section \ref{sec:results} presents two simulated studies of our method against state-of-the-art approaches and Section \ref{sec:conclusion} concludes the paper and discusses future work.

\section{Related Works} \label{sec:related}
This section presents an overview of the literature for safe planning under uncertainty and optimal experimental design.

\subsection{Safe Planning Under Uncertainty}

There are several notable methods that account for safety in the presence of uncertainty during planning.
Learning-based model predictive control \cite{aswani2013provably} decouples the planning and inference pipelines and runs them sequentially.
The inference pipeline updates system parameter estimates from sensor measurements using Bayesian inference techniques while the planning pipeline uses the parameter estimates to design a model-predictive controller that is robust to system parameter uncertainty assuming system linearity. 
Impressively, this methodology was validated via quadrotor experiments that demonstrate this approach is robust to time-varying parameters and incorrect parameter estimates \cite{bouffard2012learning} .
Other methods, such as \cite{dhiman2021control}, formulate safe controllers via control barrier functions that are computed using parameter uncertainty.
Notably, these methods can account for model, tracking, and parameter uncertainty in their controllers; however, safety is only guaranteed at discrete waypoints rather than along the entire trajectory. 

These methods leverage Bayesian inference for parameter estimation, yet they formulate data collection, and its effect on inference as a passive process.
That is they do not perform specific actions to optimally inform their parameter estimation.
As a result, these methods also implicitly assume data is easy to acquire.

\subsection{Bayesian Optimal Experimental Design}

In contrast to previous approaches, Bayesian Optimal Experimental Design (BOED) is applicable when system data is costly or time-consuming to obtain.

BOED has been shown to be useful for simultaneous localization and mapping (SLAM) \cite{sim2005global} and so-called active SLAM has seen much activity in the past decade \cite{ghaffari2019sampling, khosoussi2019reliable, ahmed2022active}.
Safety constraints have also been included in the BOED mapping paradigm \cite{teng2021toward}.
These methods are noteworthy, however they focus on data collection with respect to the robot's external environment and not on estimating uncertain parameters of a robot's dynamical model.

For the task of uncertain parameter estimation, BOED is used to learn the probability density function of a random variable $\theta$ through experimentation with as few experiments as possible by noting that different system inputs result in different parameter estimates.
Thus, BOED achieves maximal utility for scenarios where only one, or few, actions are possible.

One can quantify the amount of information gained as a function of the parameter estimates before and after a system input \cite{BEDSimulation}.
BOED is then formulated as a function of this information gain.
The amount of information gained through an experiment or action can be written as a utility function of the system input \cite{lindley1972bayesian}.
Some approaches use the estimated Kullback-Leibler divergence \cite{KLDivergence} between the posterior and prior of the parameters of interest as an information-theoretic metric \cite{huan2013simulation, LindleyBED}.
However, calculating the expected utility gained through an experiment is often intractable and requires sampling, which is time-consuming. 
Early approaches applied BOED to linear systems and formulated the optimal system input in closed-form  \cite{LinearBED, LinearBED2}; however, these methods do not extend to the non-linear regime.
More recent methods approximate information gain using Monte Carlo approximation \cite{BEDSimulation, BEDSimulation2}.

Applying BOED to robot system identification applications presents additional challenges, such as real-time operational requirements and ensuring safety.
Sampling-based approaches such as \cite{RoboticsBatch1, cooper2021bayesian} have explored BOED for robot system identification.
These BOED system identification methods, like their early counterparts, only work when the unknown system parameters of interest are modeled as Gaussian random variables.
To perform BOED for non-Gaussian systems, one must still apply sampling-based approaches.
These BOED robotic system identification approaches do not consider safety when computing optimally-informative trajectories which is essential to avoid collisions.

\section{Preliminaries} \label{sec:prelim}
This section formalizes notation used throughout the paper.
Sets and subspaces are typeset using calligraphic font.
Subscripts are primarily used as an index or to describe an particular coordinate of a vector.
Let $\R$, $\R_+$, and $\NN$ denote the spaces of real numbers, real positive numbers, and natural numbers, respectively.
Given vectors $\alpha,\beta\in\R^n$, let $[\alpha]_i$ denote the $i$-th element of $\alpha$, and let $\Int(\alpha,\beta)$ denote the $n$-dimensional box $\{\gamma \in\R^n\mid [\alpha]_i\leq[\gamma]_i\leq[\beta]_i,~\forall i=1,\ldots,n\}$.
Given arbitrary matrix $A\in\R^{n \times n}$, let $\det(A)$ be the determinant of $A$.
Given a set $\mathcal A$, denote its power set as $P(\mathcal A)$ and its cardinality as $|\mathcal A|$.
Let $\pi_{1:m}:\R^n\rightarrow\R^m$ be the projection operator that outputs the first $m$ coordinates from its argument.
Let $\rot(a)$ denote the rotation matrix computed using the $a$ coordinates.

Next, we introduce a subclass of polytopes, called zonotopes, that are used throughout this paper:
\begin{defn}
\label{def:zonotope}
A \emph{zonotope} $\Z \subset \R^n$ defined as
\begin{equation}
    \mathcal Z = \left\{x\in\R^n\mid x= c+\sum_{j=1}^\ell \beta_j g_j, \quad \beta_j \in [-1,1] \right\},
\end{equation}
with \emph{center} $c\in\R^n$ and $\ell$ \emph{generators} $g_1,\ldots,g_\ell\in\R^n$.
For convenience, we denote $\mathcal Z$ by $\zonocg{c}{G}$ where $G = [g_1, g_2, \ldots, g_\ell ]\in\R^{n\times\ell}$
\end{defn}

We define the multiplication of a matrix $A$ of appropriate size with a zonotope $\Z$ as
\begin{equation}
\label{eq: zono-matrix mult}
   A \Z = \left\{x\in\R^n\mid x= A c+\sum_{j=1}^\ell \beta_j A g_j, ~ \beta_j \in [-1,1] \right\}.
\end{equation}
Note that $A \Z$ is equal to the zonotope $\zonocg{A c}{A G}$. 

\subsection{Probability}

Next, we review several relevant results from probability theory.

\begin{defn}  [{\cite{smith2013uncertainty}}]
    A probability space $(\Omega, \mathcal{F}, P)$ is comprised of a sample space $\Omega$, a $\sigma$-algebra $\mathcal{F}$ of subsets of $\Omega$, and  a probability function $P:\mathcal{F}\rightarrow [0,1]$.
\end{defn}
\noindent Before defining a probability density function, we introduce random variables:
\begin{defn}
    A continuous random variable is a function $X: \Omega \rightarrow \R$ with a cumulative distribution function $F_X:\R \rightarrow [0,1]$ such that

    \begin{equation}
        F_X(x) = P\{\omega \in \Omega \mid X(\omega) \leq x\},
    \end{equation}

    \noindent where $x=X(\omega)$ is termed the \textit{realisation} of the random variable for some event $\omega \in \Omega$.
\end{defn}
\noindent Given the cumulative distribution function for a random variable, we can define the probability density function for a random variable:
\begin{defn}
    The probability density function $p_X$ of random variable $X$ is defined as the derivative of the cumulative density function of $X$, or
    \begin{equation}
        F_X(x) = \int_{-\infty}^x p_X(s) \mathrm{d}s.
    \end{equation}

    \noindent Note, we omit the subscript of the probability density function for the remainder of this paper for notational convenience.
\end{defn}

Lastly, we present Bayes' theorem relating conditional probabilities.
This theorem is used throughout the remainder of this paper.
\begin{thm} \label{eq:bayes}
    Given two random variables $A$ and $B$ conditioned on one another, Bayes' theorem states that the following equality hold
    \begin{equation}
        P(A \mid B) = \frac{P(B \mid A) P(A)}{P(B)}.
    \end{equation}
\end{thm}


For notational convenience we use a semicolon to denote dependence of a probability density function on a deterministic value.
For example the probability density function of a random variable $x$ dependent on a value $k$ is denoted as $p(x ; k)$, whereas the probability density function of a random variable $x$ conditioned on another random variable $z$ is denoted as $p(x \mid z)$.

\subsection{Cubature} \label{subsec:cubature}

Quadrature is a method of numerically approximating one-dimensional integrals as a weighted summation of the integrand.
Cubature is the generalization of quadrature to multi-dimensional integrals.
We formalize cubature as follows:

\begin{defn} \label{def:cubature}
Given an integrable function $f:\R^n \rightarrow \R$, and domain $\X \subset \R^n$, \emph{cubature} is any method to approximate the integral of $f$ as a weighted sum of function evaluations:

\begin{equation}
    \int_{\X} f(x) \mathrm{d}x \approx \sum_{j=1}^N w_j f(x_j),
\end{equation}

\noindent where $w_j \in \R$ and $x_j \in \R^n$ are pre-computed \emph{cubature weights} and \emph{points}, respectively.
These are computed as a function of the integration domain $\X$ and the number of desired cubature points $N$ and do not depend on the integrand $f$.
\end{defn}

Cubature formulations exist for both bounded and unbounded integration domains, with each formulation generating different sets of cubature weights and points.
Common cubature methods are Gauss-Legendre \cite{swarztrauber2003computing}, Gauss-Hermite \cite{liu1994note}, and Clenshaw-Curtis \cite{gentleman1972implementing}.
This paper uses the Clenshaw-Curtis cubature method to approximate integrals over bounded integral domains; however, one may use Gauss-Hermite or Gauss-Laguerre cubature to approximate integrals over unbounded integral domains.

The accuracy of cubature methods is dependent on the the integrand.
The Gauss-Hermite cubature is applicable for probability distributions belonging to the exponential family while Clenshaw-Curtis cubature is applicable to probability distributions with bounded sample space.

\section{Robot Model, Environment, and Sensing Model} \label{sec:model}
This section introduces the parameterized robot dynamics (Section \ref{subsec:robot_model}) and measurement model (Section \ref{subsec:env_and_sensing}) that are later used to compute optimally-informative system trajectories.
These definitions allow us to formalize the definition of safety for the robot (Section \ref{subsec:safety}).

\subsection{Robot Dynamics Model and Trajectory Parameterization} \label{subsec:robot_model}

Given a dynamic system with states $x(t) \in \R^n$, system inputs $u(t) \in \R^m$, and time-invariant system parameters $\theta \in \Theta \subset \R^p$, the \emph{robot dynamics} are

\begin{equation} \label{eq:dyn}
    \dot{x}(t; u(t), \theta) = f(x(t;u(t),\theta), u(t), \theta).
\end{equation}

We make a mild assumption about the parameter sample space which we will later use to bound the possible future states of the system.

\begin{assum} \label{assum:bounded}
The parameter sample space $\Theta$ is connected and bounded and can be represented as $\Theta = \Int(\underline{\theta},\overline{\theta})$ where $\underline \theta$ and $\overline \theta$ are the lower and upper bounds on the system parameters, respectively.
\end{assum}

In Section \ref{sec:bed}, we wish to compute an input to our system that optimally reduces our uncertainty about the parameters.
To make this computationally tractable, we parameterize the set of control inputs.
Before we describe this process, we start by defining how control inputs are applied to our system:

\begin{assum} \label{assum:tplan}
During each planning iteration starting from time $\tz$, the agent has $\tplan$ seconds to find a control input.
This control input is applied during the time interval $\mathcal{T} = [\tz + \tplan, \tz + \tplan + \tf]$ where $\tf \geq \tplan$ is specified by the user. 
In addition, the state of the agent at time $\tz + \tplan$ is known. 
\end{assum}

Next, we define the parameterized system input: 

\begin{assum} \label{assum:system_input}
Let $\K$ denote the set of \emph{trajectory parameters}.
Let the parameterized state input be any square integrable function, $u: \mathcal{T} \times \K \to \R^m$ that satisfies the following properties:
\begin{enumerate}
\item For all $k\in\K$, there exists a time instant $\tnb < \tz + \tplan + \tf$, such that for $t > \tnb$ the input $u(t,k)$ brings the agent to a stop before $\tz + \tplan + \tf$ and the agent remains stopped;
\item $u$ is continuously differentiable with respect to its second argument $k$.
\end{enumerate}
\end{assum}

\noindent The first property ensures that any trajectory parameter computed by the optimization problem can keep the robot persistently safe.
The second property ensures that we can apply optimization to quickly solve the trajectory optimization problem through gradient descent.

\begin{rem} \label{rem:implicit_motions}
Parameterizing the trajectory also allows us to implicitly encode behaviours into the motion of the robot.
Examples of this are shown in Section \ref{sec:results}.
\end{rem}

Next, we define the footprint of the robotic agent, or the space it takes up in the world.
We begin by introducing the world space:
\begin{defn}
    Let $\W \subset \R^w$ be the \emph{world space} and let the first $w \in \NN$ coordinates of $x(t;k,\theta)$ correspond to the world space, or $ \pi_{1:w}(x(t; k, \theta)) \in \W$.
\end{defn}

\noindent Next, we define the agent footprint:
\begin{defn}
    The agent is a rigid body that can be described using a zonotope $\Oego$ such that $\Oego \subset \W$.
    $\Oego$ is called the agent \emph{footprint}.
\end{defn}
\noindent For arbitrary time $t$, given the state $x(t;k,\theta)$ of the agent that starts from initial condition $x_0 \in \R^n$, system parameters $\theta \in \Theta$, and applies a control input parameterized by $k\in\K$, the agent's \emph{forward occupancy} at time $t$ is
\begin{equation}
\begin{split}
    \hspace{-0.2cm}\E\big(t;x_0,k,\theta\big) := \rot(x(t; k, \theta))\cdot\Oego + \\ + \pi_{1:w}(x(t; k, \theta)),
    \end{split}
\end{equation}
which is a zonotope by \eqref{eq: zono-matrix mult}.

\subsection{Environment and Sensing Model} \label{subsec:env_and_sensing}

Next, we define the environment of the agent:
\begin{defn}
An \emph{obstacle} is a set $\Oobs(t)\subset \W$ that the agent should not collide with at time $t$, where $i\in\I$ is the index of the obstacle and $\I$ has a finite number of elements.
Additionally, the set of all obstacles at time $t$ is denoted as $\Oobsset(t) = \cup_{i \in \I} \Oobs(t)$.
\end{defn}

\noindent We make the following assumption about obstacles in the agent's surroundings.
\begin{assum}
    The agent can detect all obstacles that it is able to collide with in the interval $[\tz + \tplan, \tz + \tplan + \tf]$.
    The agent has this information at time $\tz$.
\end{assum}

Independent of obstacle detection, the agent obtains system measurements as a function of its states at time $t$, though these measurements may be corrupted by additive Gaussian noise:
\begin{equation} \label{eq:measurement}
    y(t; k, \theta) = g(x(t;k,\theta)) + \omega.
\end{equation}
\noindent where $y(t; k, \theta) \in \R^d$, $\omega \sim \Gauss(0, \Sigma)$, and $\Sigma \in \R^{d \times d}$ is the measurement noise covariance.

\subsection{Safety} \label{subsec:safety}

Given the robot's forward occupancy, $\E\big(t;x_0,k,\theta\big)$, and the set of obstacles, $\Oobsset(t) = \cup_{i \in \I}\Oobs(t)$, at time $t$, we formalize not-at-fault behaviour as follows and use this definition to define safety.
\begin{defn} \label{def:safety}
A maneuver is \emph{not-at-fault} if the agent is stopped, or if it is never in collision with any obstacle while it is moving, i.e., 
\begin{equation} \label{eq:not_at_fault_safety}
    \E \big(t;x_0,k,\theta \big) \cap \Oobsset(t) = \emptyset  \quad \forall t \in \mathcal{T},
\end{equation}
for all $t$ where the agent is moving. 
\end{defn}
\noindent From \cite{vaskov2019towards}, formulating not-at-fault safety as \eqref{eq:not_at_fault_safety} ensures the agent's forward occupancy never intersects any obstacle over the entire time horizon, thus ensuring the agent is never in collision while it is in motion.

We note that, in contrast to active SLAM methods which assume uncertainty with respect to the robot's environment, we instead assume uncertainty with respect to the system parameters of the robot dynamics and assume an environmental representation is given.
Accounting for both the uncertainty in the robot's environment and dynamic model parameters is out-of-scope for this paper.


\section{Bayesian Experimental Design Theory} \label{sec:bed}
This section reviews the definition of information gain, its relation to probability theory, and introduces a cost function to maximize expected information gain given a system trajectory.
The section concludes by formulating an optimization problem to compute maximally-informative, safe trajectories.

\subsection{Expected Information Gain} \label{subsec:info_gain}

Before introducing how to compute expected information gain, we first introduce the functions used in this computation. 
We obtain data about our system through state measurements $y$ as in \eqref{eq:measurement}, however, we only receive measurements at discrete time steps.
To describe this process, pick a time step $\Delta_t \in \R_{+}$ such that $t_f / \Delta_t \in \NN$ and partition the time interval $[0, t_f]$ into $t_f / \Delta_t$ time segments as $T_j = [(j-1)\Delta_t, j\Delta_t]$ for each $j\in \mathcal{J} = \{1,2,\ldots,t_f/\Delta_t\}$.
Then we denote a measurement at time $t_j := j\Delta_t$ as $y(t_j; k, \theta) := y_j(k, \theta)$. 

Measurements in \eqref{eq:measurement} are functions of the system state and additive Gaussian white noise.
Thus, the measurement is also a Gaussian random variable with \emph{measurement likelihood}

\begin{equation} \label{eq:likelihood}
    p(y_j \mid \theta; k) = \N(g(x(t_j;k, \theta), \Sigma).
\end{equation}

The measurement marginal distribution, also called the \emph{evidence}, is computed using the measurement likelihood.

\begin{equation} \label{eq:evidence}
    p(y_j ; k) = \int\limits_\Theta p(y_j \mid \theta; k) p(\theta) \mathrm{d}\theta.
\end{equation}

We also introduce the prior and posterior probability estimates for the unknown system parameters $\theta$.
The \emph{prior}, $p(\theta)$, represents our belief in the distribution of $\theta$ at the beginning of planning, while the \emph{posterior}, $p(\theta \mid y_j; k)$, represents our belief in the distribution of $\theta$ after we have executed a plan and collected data.
Bayes' Theorem \eqref{eq:bayes} is used to compute the posterior as a function of the prior, measurement likelihood, and measurement marginal:

\begin{equation} \label{eq:posterior}
    p(\theta \mid y_j ; k) = \frac{p(y_j \mid \theta; k) p(\theta)}{p(y_j; k)}.
\end{equation}

Given \eqref{eq:posterior}, a question naturally arises: How does the trajectory parameter $k$ affect the posterior?
To answer this question, we first define information gain and show that it is a metric for computing optimally-informative actions.
\begin{defn}[\cite{lindley1972bayesian}] \label{def:info_gain}
    \emph{Information gain} is the amount of information obtained through an observation about a random variable, in this case, the unknown system parameters $\theta$.
    In particular, the information gain, $L$, is the Kullbeck-Leibler divergence between the posterior and prior of the random variable:
    \begin{align} \label{eq:kldiv}
        L(k, y_j) &:= D_{KL}\left[p(\theta \mid  y_j; k) \parallel p(\theta) \right] \\
        &= \int\limits_\Theta \ln \left(\frac{p(\theta \mid y_j; k)}{p(\theta)} \right) p(\theta \mid y_j; k) \mathrm{d}\theta.\label{eq:theta_posterior_info_gain}
    \end{align}
\end{defn}

\noindent From \eqref{eq:posterior} the posterior of $\theta$ is dependent on the trajectory parameter $k$, therefore the information gain is a function of $k$.
Additionally, the information gain is also a function of the random variable $y_j$.
Thus, motivated by \cite{lindley1972bayesian}, we instead compute the \emph{expected information gain} over the measurements $y_j$

\begin{align}
J(k, t_j) &:= \EX_{y_j, \theta} [L(k, y_j)]  \\
&= \int\limits_{\R^d} \int\limits_\Theta L(k, y_j) p(y_j, \theta ;  k) \mathrm{d}\theta \mathrm{d}y_j, \label{eq:bed_short}
\end{align}

\noindent such that $\R^d$ is the sample space of $y_j$ given that $y_j$ is a Gaussian random variable.

For a prior on $\theta$ described by an arbitrary probability distribution, computing the posterior of $\theta$ is challenging and may not have a closed-form solution.
To avoid this challenge, we refactor \eqref{eq:bed_short} by applying  Bayes' Theorem such that it is a function of only the measurement likelihood, measurement marginal, and parameter prior:

\begin{align} \label{eq:eig_measurements}
    J(k,t_j) &= \int\limits_{\R^d} \int\limits_\Theta L(k, y_j)p(y_j, \theta ;  k) \mathrm{d}\theta \mathrm{d}y_j \\
    &= \int\limits_{\Theta} D_{KL}\big[p(y_j \mid \theta; k) \parallel p(y_j ; k) \big] p(\theta) \mathrm{d}\theta. \label{eq:eig}
\end{align}

Lastly, \eqref{eq:eig_measurements} only considers measurements at a given time $t_j$.
To account for measurements taken over the entire trajectory, we sum over the time interval $T = [t_1,\dots,t_{|\J|}]$.

\begin{equation} \label{eq:eig_final}
    J(k) = \sum_{t_j \in T} J(k, t_j).
\end{equation}


\subsection{Formulating the Optimization Problem}

We compute safe, optimally-informative system trajectories parameterized by $k$, as per Section \ref{subsec:robot_model}, using the expected information gain metric defined in Section \ref{subsec:info_gain} as the cost function using the following optimization problem

\begin{align*}
    \min_{k \in \K} & ~ J(k) \hspace{5.3cm} \opt\\
    \text{s.t.} 
    & ~ \E \big(t;x_0,k,\theta \big) \cap \Oobs(t) = \emptyset \quad \forall t \in \mathcal{T},
\end{align*}
 where the safety constraint is defined in Definition \ref{def:safety}.
Unfortunately, for arbitrary prior probability distributions, computing the measurement marginal \eqref{eq:evidence} or the expected information gain \eqref{eq:eig} in closed-form is challenging.
This makes solving $\opt$ difficult.
In the proceeding section we compute a differentiable approximation of the expected information gain function and a differentiable over-approximation of the safety constraint.
We then use these approximations to formulate a computationally tractable approximation of $\opt$.

\section{Safe, Differentiable Approximation to \opt} \label{sec:alg}
This section describes how to approximate the expected information gain \eqref{eq:eig_final} such that the approximation is differentiable with respect to the trajectory parameter $k$.
Additionally, this section introduces an over-approximation for the safety constraint and shows that this over-approximation is also differentiable with respect to $k$.
This section concludes by presenting a tractable relaxation of $\opt$ and an algorithm for computing optimal trajectory parameters for dynamical systems.

\subsection{Approximating the Cost Function} \label{subsec:relax_cost}

We start by presenting an approximation of the expected information gain function using cubature.
This approximation makes no assumptions on the form of the prior, $p(\theta)$, and only requires the measurement likelihood to be Gaussian as in \eqref{eq:likelihood}.
Approximating \eqref{eq:eig_final} first requires an approximation of the measurement marginal as is described in the following proposition whose proof can be found in Appendix A.

\begin{prop} \label{prop:marginal}
By Definition \ref{def:cubature}, the measurement marginal is approximated using cubature such that
\begin{align} \label{eq:marginal}
    p(y_j ; k) &= \int\limits_\Theta p(y_j \mid \theta; k) p(\theta) \mathrm{d}\theta\\
    &= \sum_{i=1}^N \alpha_i p(y_j \mid \theta_i ; k),
\end{align}
\noindent where $w_i$ and $\theta_i$ are the cubature weights and points, respectively, and
\begin{equation}
     \alpha_i = \frac{w_i p(\theta_i)}{\sum_{l=1}^N w_l p(\theta_l)}.
\end{equation}
\end{prop}

\begin{rem}
    Proposition \ref{prop:marginal} indicates that, given a Gaussian likelihood, the measurement marginal can be approximated as a Gaussian Mixture Model (GMM).
    More accurate approximations of the marginal can be computed using more cubature points.
\end{rem}

We use this approximation of the marginal distribution to compute \eqref{eq:eig_measurements} and describe a differentiable approximation of the expected information gain using the following Corollary:

\begin{cor} \label{cor:eig_cubature}
Let $x_0 \in \R^n$ be the initial state of the system, $y_j(k, \theta) \in \R^d$, and let $p(y_j \mid \theta ; k) \sim \N(g(x(t; k, \theta)), \Sigma)$ be the measurement likelihood. 
If the marginal is represented as a Gaussian Mixture Model as in Proposition \ref{prop:marginal}, then the expected information gain \eqref{eq:eig} is

\begin{equation}\label{eq:eig_approx}
    \Tilde{J}(k, t_j) = \sum_{i=1}^N w_i p(\theta_i) \left( c - \ln \sum_{l=1}^N \alpha_l Z_{i,l} \right),
\end{equation}
\noindent where $w_i$ and $\theta_i$ are the cubature weights and points of the parameter sample space $\Theta$ and
\begin{align}
    c &= -\frac{1}{2} \left( \ln\left((2 \pi)^d \det(\Sigma)\right) + d \right) \\
    Z_{i,l} &= ((2 \pi)^d \det(2\Sigma))^{-1/2} \cdot \\
        & \hspace*{1cm} \cdot \exp\left(-\frac{1}{2} (\mu_i - \mu_l)^{\intercal} (2\Sigma)^{-1} (\mu_i - \mu_l)\right) \nonumber \\
    \mu_i &= g(x(t_j; k, \theta_i)).
\end{align}
\end{cor}

\noindent A proof is provided in Appendix B.

If we sum over the time interval $T = [t_1, \dots, t_{|\J|}]$ to obtain an approximation of \eqref{eq:eig_final}.
\begin{equation}
    \Tilde{J}(k) = \sum_{t_j \in T} \Tilde{J}(k, t_j).
\end{equation}

\subsection{Over-Approximating the Safety Constraint} \label{subsec:relax_constraint}

We now show how to over-approximate the safety constraint in $\opt$ by over-approximating the robot's forward occupancy.
Computing $\E\big(t;x_0,k,\theta \big)$ exactly is difficult unless one assumes a special form for the robot dynamics (\emph{e.g.}, linear).
By Assumption \ref{assum:bounded} the system parameter sample space is bounded, thus we compute an over-approximation of $\E\big(t;x_0,k,\theta \big)$ as follows:
\begin{assum} \label{assum:reachable_sets}
    Let $x$ be a solution to \eqref{eq:dyn} starting from initial condition $x_0 \in \R^n$ with trajectory parameter $k\in\K$ and $\theta \in \Theta$.
    For each $j\in\J$, there exists a map $\xi_j: \R^n \times \K \rightarrow P(\W)$ such that 
    \begin{enumerate}
        \item $\xi_j(x_0,k)$ contains the agent's footprint during $T_j$, i.e. $\cup_{t\in T_j} \E(t;x_0,k,\theta)\subseteq \xi_j(x_0,k)$
        \item $\xi_j(x_0,k)$ is a zonotope of the form $\zonocg{c_j(x_0)+A_j k}{G_j}$ with some linear function $c_j:\R^n \rightarrow \R^n$, some matrix $A_j\in\R^{n\times n_p}$ and some $n$-row matrix $G_j$.
        \item $\xi_j(x_0,k)$ accounts for all possible uncertain system parameters $\theta \in \Theta$.
    \end{enumerate}
Let $\xi_j(x_0, k)$ denote the \textit{reachable set} of the robot for the time interval $T_j$.
\end{assum}

There exists various techniques to generate this reachable set representation for a numerous robotic systems \cite{kousik2019_quad_RTD,liu2022refine,michaux2023}.
Next, we make the following assumption about dynamic obstacles within the environment motivated by \cite{vaskov2019towards}.

\begin{assum} \label{assum:obstacle_intervals}
    We over-approximate an obstacle within the time interval $T_j$ using a zonotope $\Oobs(T_j)$ such that $\cup_{t \in T_j} \Oobs(t) \subset \Oobs(T_j)$.
    We denote the union of all such obstacle representations over the time interval $T_j$ as $\Oobsset(T_j) = \cup_{i \in \I} \Oobs(T_j)$.
\end{assum}

We use the following theorem to relate this over-approximation of obstacles and the agent's footprint to the over-approximation of the safety constraint:

\begin{thm} [{\cite[Thm 23]{liu2022refine}}]
From Assumptions \ref{assum:reachable_sets} and \ref{assum:obstacle_intervals} we obtain the agent's forward reachable sets and the zonotope representation of obstacles over the time interval $T_j$.
The collision set is then over-approximated using these representations as
\begin{equation}
    \E(t; x_0,k,\theta) \cap \Oobs(t) \subset \E_j(x_0,k) \cap \Oobsset(T_j)
\end{equation}
for $t \in T_j$.
Using the over-approximative collision set, we over-approximate the constraint $\E_j(x_0,k) \cap \Oobsset(T_j) = \emptyset$.
The over-approximative safety constraint is linear with respect to the trajectory parameter $k$ and thus the gradient with respect to $k$ is computable.
\end{thm}

\subsection{An Implementable BOED}

Combining the results of Sections \ref{subsec:relax_cost} and \ref{subsec:relax_constraint} we present a differentiable approximation to $\opt$ where both the cost and constraint are differentiable with respect to $k$:
\begin{align*}
    \min_{k \in \K} & ~ \Tilde{J}(k) \hspace{5.3cm} \optE\\
    \text{s.t.} 
    & ~ \E_j(x_0,k) \cap \mathcal{O}(T_j) = \emptyset \quad \forall j \in \J.
\end{align*}

The algorithm to compute safe, optimally-informative trajectories is shown in Algorithm 1.
The reachable sets $\{\xi_j\}_{j\in\J}$ are computed offline using CORA Toolbox \cite{althoff2018b}.
These reachable sets, along with an initial condition $x_0 \in \R^n$ and an initial trajectory parameter estimate $k_0 \in \K$, are specified at runtime.
We assume the agent is either initially stationary or is implementing a previously computed safe, optimal trajectory.

At time $\tz$, obstacles are first detected within the sensor range (Line \ref{alg:line:obstacles}).
We use Clenshaw-Curtis cubature on the parameter space $\Theta$ and compute the cubature weights and points used for the expected information gain computation (Line \ref{alg:line:quadrature}).
The optimal trajectory parameter $k^{*}$ is computed using our relaxed optimization formulation $\optE$ (Line \ref{alg:line:opt}), and this parameter is then sent to the controller to track the desired optimally-informative trajectory.
If an optimal trajectory cannot be computed before $\tplan$ (see Assumption \ref{assum:system_input}), the agent begins the braking maneuver of the previous optimal trajectory (Line \ref{alg:line:braking}).
If the agent was initially stationary and no optimal trajectory can be computed before $\tplan$, the agent remains stationary.
Thus, the following Lemma holds.
\begin{lem}
\label{lem:recursive_safety}
Beginning at rest, if the agent applies any feasible solution, $k^*\in \K$, of \optE{}, then the agent is not-at-fault during $[0,\tf]$.
\end{lem}

\begin{algorithm}[t] \label{alg:bed}
    \caption{Safe Bayesian Experimental Design}
    \begin{algorithmic}[1]
        \REQUIRE $x_0 \in \R^m$, $\Theta \subset \R^p$, $k_0 \in \K$,$\{\xi_j\}_{j\in\J}$
        \STATE \textbf{Initialize:} $\kstar = k_0$, $t = 0$
        \STATE \textbf{Loop:}
            \STATE \quad $\{\Oobs(T_j)\}_{(j,i) \in \J \times \I}  \leftarrow \texttt{SenseObstacles}()$ \label{alg:line:obstacles}
            \STATE \quad $\{w_l, \theta_l\}_1^N \leftarrow \texttt{Quadrature}(\Theta, N)$  \label{alg:line:quadrature}
            \STATE \quad \textbf{Try:} $\kstar \leftarrow \texttt{Opt-E}(x_0, w_l, \theta_l,\{\Oobs(T_j)\}_{(j,i) \in \J \times \I})$ \label{alg:line:opt}
            \STATE \quad \textbf{Catch:} $\tplan \geq \tf$, then $\textbf{break}$ \label{alg:line:braking}
        \STATE \textbf{End}
        \STATE \textbf{Execute:} $\kstar$ during $[0, \tf]$

    \end{algorithmic}
\end{algorithm}

\section{Experiments} \label{sec:results}
We compare our formulation for safe, differentiable Bayesian Optimal Experimental Design against state-of-the-art methods from the literature.
We start by demonstrating our approach on the non-linear measurement function introduced in \cite{huan2013simulation} and compare our method with the typical Monte Carlo integration technique.
Next, we apply our formulation to an autonomous vehicle with a high-fidelity dynamical model and compare against a baseline sampling-based BOED approach \cite{cooper2021bayesian}.
We demonstrate the differentiability of our BOED approximation enables computation of a maximally-informative input without needing to sample over the entire system input domain.
Additionally, we illustrate our method is able to compute safe, optimally-informative system trajectories for a single planning iteration, exemplifying the utility of BOED for situations where only one, or few, actions are taken.
Experiments were conducted on a laptop with an 2.5GHz i7-11900H CPU, RTX 3080 Mobile, and 32 GB of memory.

\subsection{Nonlinear Measurement Function}

Consider the non-linear measurement model presented in \cite{huan2013simulation}

\begin{equation} \label{eq:nonlin_model}
    y(\theta, u) = \theta^3 u^2 + \exp(-|0.2 - u|) + \omega,
\end{equation}

\noindent where $\omega \sim \N(0, 10^{-4})$ is additive Gaussian noise and the prior is $p(\theta) = U(0,1)$.

The goal is to compute the input $u \in [0,1]$ which minimizes the entropy of the posterior of $\theta$.
This corresponds to computing the input that maximizes the expected information gain from \eqref{eq:eig}.
For this example, we need not consider system state evolution, trajectory parameterization, nor safety.
Because the system input domain is compact and bounded we do not parameterize it and therefore we reformulate $\optE$ as
\begin{equation} \label{eq:xun_approx}
    u^{*} = \max_{u \in [0,1]} \Tilde{J}(u),
\end{equation}
\noindent without the safety constraint.

\begin{figure} [!t]
    \centering
    \includegraphics[width=\columnwidth]{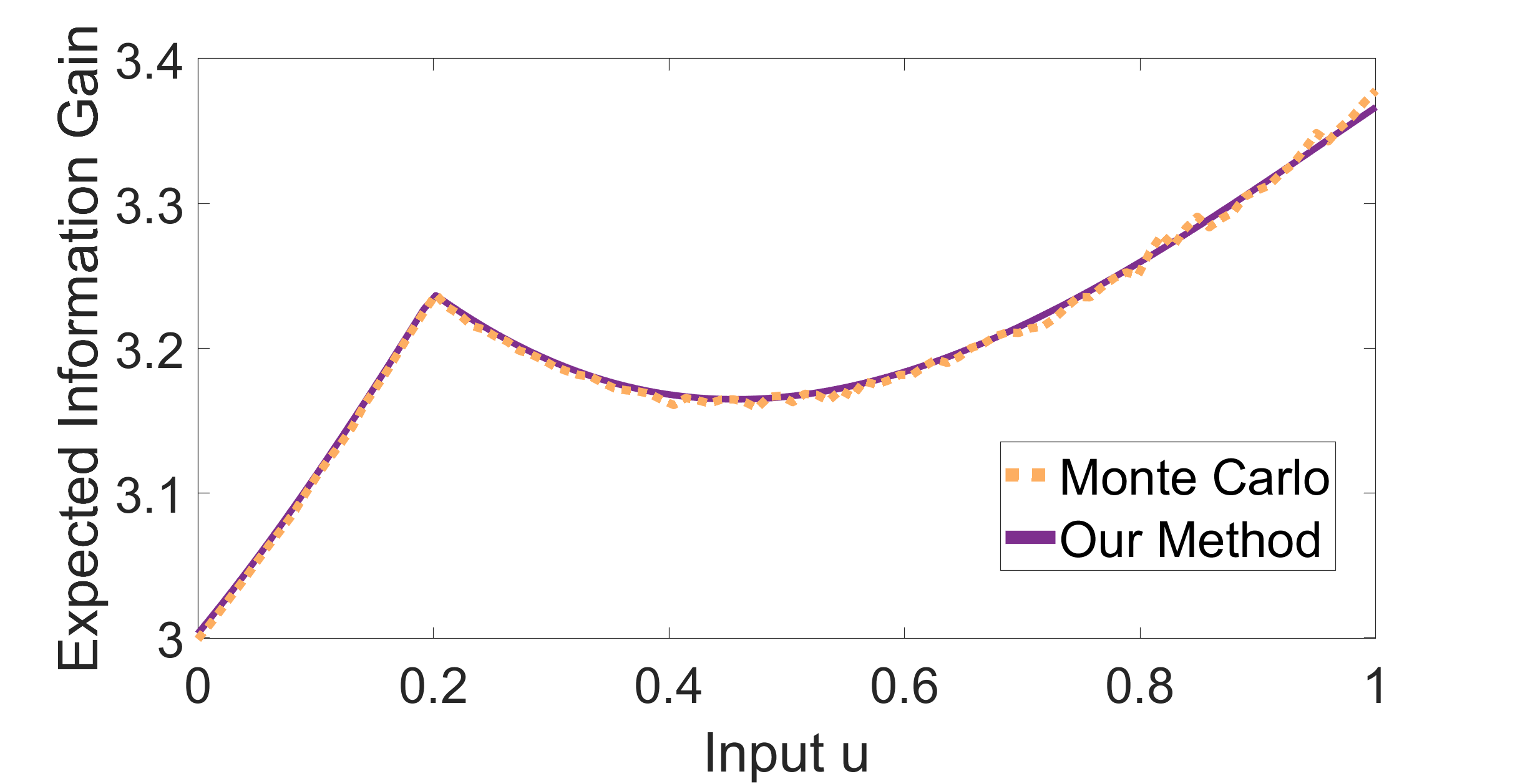}
    \caption{Expected Information Gain approximated using Monte Carlo \cite{huan2013simulation} and our approach for the non-linear model \eqref{eq:nonlin_model}. Our differentiable approximation achieves accurate expected information gain estimates with only $100$ cubature points while the Monte Carlo estimate requires $10^5$ samples. Using gradient ascent, an optimal input of $u=0.2$ or $u=1.0$ is found depending on the initial condition.}
    \label{fig:xun_example}
\end{figure}

Figure \ref{fig:xun_example} illustrates the estimated expected information gain using Monte Carlo estimation \cite{huan2013simulation}, shown in orange, using $10^5$ samples alongside our differentiable approximation of the expected information gain computed using \eqref{eq:xun_approx}, shown in purple.
We only use $100$ cubature points to compute \eqref{eq:eig_approx}.
This example shows that our differentiable formulation is able to closely approximate the expected information gain using orders of magnitude fewer samples than Monte Carlo estimation.

\subsection{Autonomous Vehicle Model}

We adapt the high-fidelity, front-wheel drive vehicle model from \cite{cooper2021bayesian, liu2022refine} for this experiment.
The vehicle's states at time $t$ are $x(t) = [x_1(t), x_2(t), x_3(t), x_4(t), x_5(t), x_6(t)]$ corresponding to longitudinal ($x_1(t)$) and lateral ($x_2(t)$) position of the vehicle in the world frame, yaw ($x_3(t)$) in the world frame, longitudinal ($x_4(t)$) and lateral ($x_5(t)$) velocity in the vehicle's body frame, and yaw rate of change ($x_6(t)$).

The high-speed, high-fidelity dynamics for this vehicle are
\begin{align} \label{eq:dynamics_hi}
    \dot{x}^{hi}(t) & = 
    \begin{bmatrix}
        x_4(t)\cos x_3(t) - x_5(t)\sin x_3(t)\\ x_4(t)\sin x_3(t) + x_5(t)\cos x_3(t) \\ x_6(t)\\
        \frac{1}{m}\big(F_\text{xf}(t;k)+F_\text{xr}(t)\big) + x_5(t)x_6(t) \\ 
        \frac{1}{m}\big(F_\text{yf}(t;k)+F_\text{yr}(t)\big) - x_4(t)x_6(t)\\
        \frac{1}{I_\text{zz}} \big(l_\text{f} F_\text{yf}(t;k) - l_\text{r} F_\text{yr}(t)\big)
    \end{bmatrix},
\end{align}

\noindent where $m$ is the vehicle's mass, $I_{zz}$ is the vehicle's moment of inertia, $l_f$ and $l_r$ are the distances from the vehicle's center of mass to the front and rear tire centers, respectively, and $F_{ij}(t;k)$ are the tire forces which are the inputs to the system.
See Appendix C for a description of how to compute the tire forces.

When the vehicle's longitudinal speed is less than $5\text{m/s}$ we use the following low-speed lateral velocity and yaw rate dynamics:
\begin{align}
    \dot{x}_5^{low}(t) &= l_r x_6^{low}(t) - \frac{m l_f}{c_{\alpha r} l} x_4^2(t) x_6^{low}(t) \\
    \dot{x}_6^{low}(t) &= \frac{\delta x_4(t)}{l + C_{us} x_4(t)^2},
\end{align}
\noindent where $l = l_f + l_r$, $c_{\alpha i}$ is the cornering stiffness, $\delta$ is the steering angle, and 
\begin{equation}
    C_{us} = \frac{m}{l} \left( \frac{l_r}{c_{\alpha f}} - \frac{l_f}{c_{\alpha r}} \right).
\end{equation}

As in \cite{cooper2021bayesian} we only observe the longitudinal and lateral velocities of the vehicle through noisy measurements, where the measurement equation is given as:

\begin{equation}
    y(t_j; k, \theta) = \begin{bmatrix}
        0 & 0 & 0 & 1 & 0 & 0 \\
        0 & 0 & 0 & 0 & 1 & 0
    \end{bmatrix}
    x(t_j; k, \theta) + \omega,
\end{equation}

\noindent where $\omega \sim \N(0, 0.01 \cdot \mathbb{I})$ and $\mathbb{I} \in \R^{2 \times 2}$ is the identity matrix.

The uncertain system parameters are $\theta = [m, l_f, l_r]$.
We chose a prior $\theta_i \sim U(0.95 \hat{\theta}_i, 1.05 \hat{\theta}_i)$ which corresponds to the $\pm 5\%$ interval centered at the parameter means $\hat{\theta}_i$.
The parameter means are $\hat{m} = 1500 \text{kg}$, $\hat{l}_f = 1.13$m, and $\hat{l}_r = 1.67$m.
The values for the other parameters are given in Appendix D.
The vehicle footprint at $t_0$ is chosen as $\mathcal O^{ego}:=\Int([-1.2,-0.55]^T,[1.2,0.55]^T )\subset\W$ which corresponds to a vehicle length of 2.4m and a width of 1.1m.
Reachable sets $\{\xi_j\}_{j\in\J}$ satisfying Assumption \ref{assum:reachable_sets} are computed using Lemma 21 of \cite{liu2022refine}.

\begin{figure} [!t]
    \centering
    \includegraphics[width=\columnwidth]{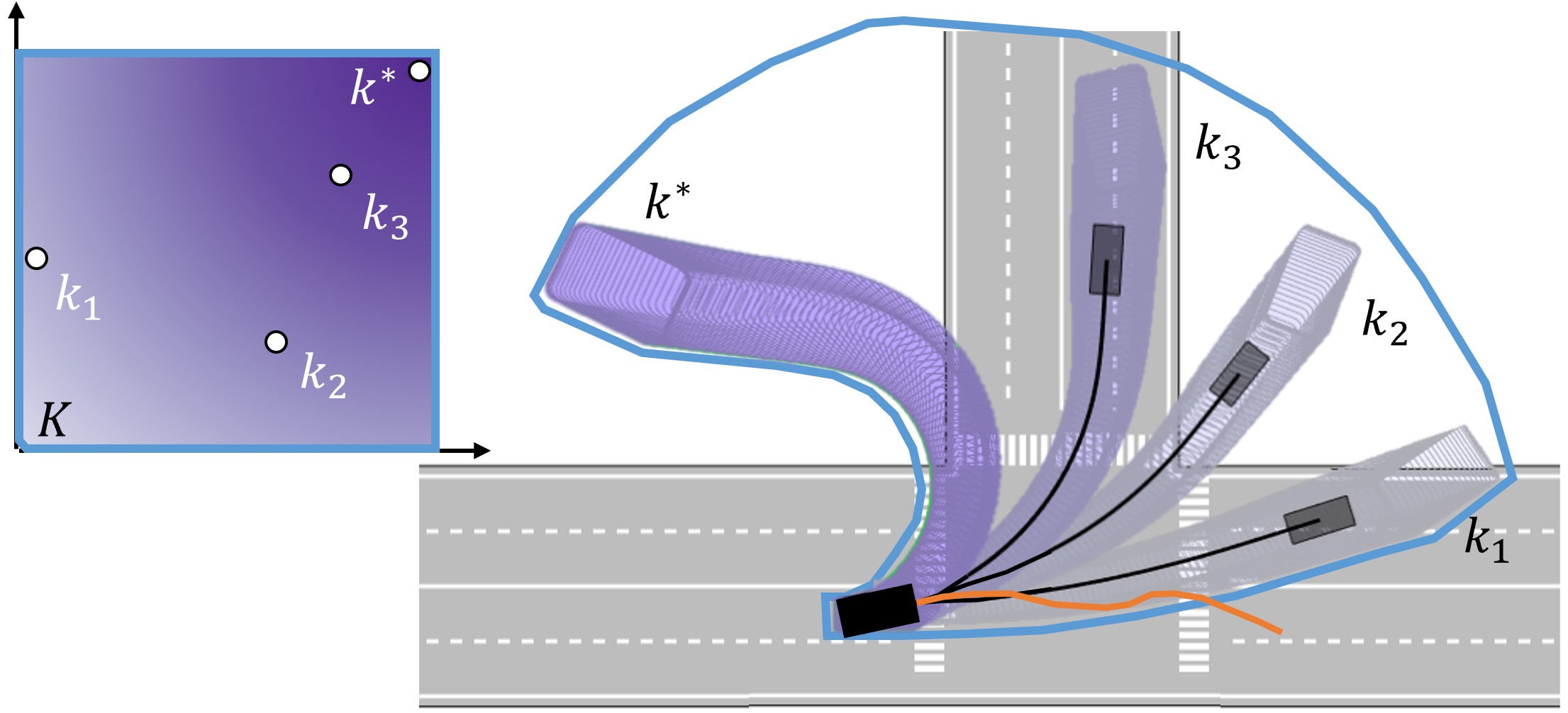}
    \caption{The optimally-informative trajectory computed using $\optE$ for a turning maneuver without considering the safety constraint. The uncertain parameters of interest appear prominently in the yaw rate of change within the vehicle's dynamics \eqref{eq:dynamics_hi}, thus to maximize expected information gain with respect to these parameters the chosen trajectory turns as much as possible. The union of the forward-reachable sets for the entire parameter space is shown in blue, the forward reachable sets for the optimal parameter $k^{*}=[10,1]$, as well as for other parameter samples $k_1$, $k_2$, and $k_3$ are shown and color-coded depending on their expected information gain, where a darker color indicates a higher value.
    The orange line corresponds to the optimally-informative trajectory computed using the baseline \cite{cooper2021bayesian}.}
    \label{fig:rover_example}
\end{figure}

We compare our method against that of \cite{cooper2021bayesian}, which computes the expected information gain for a set of sampled inputs by assuming a Gaussian prior on $\theta$ and then applying the Dual Unscented Kalman Filter \cite{davoodabadi2014identification} to estimate the posterior of $\theta$ for each input sample.
Given that the prior and posterior estimates are Gaussian, \eqref{eq:eig_final} has a closed-form solution and the optimally-informative input sample is chosen at each time step.
As \cite{cooper2021bayesian} does not use parameterized inputs, we modify the dynamics such that the inputs are the tire rotation speed, $\omega$ controlling the longitudinal tire force $F_{xf}(t)$ and steering angle $\delta$.
See Appendix E for how $F_{xf}(t)$ is computed using the tire rotation speed.
 
\subsubsection{Without Safety}

For this comparison, we do not consider safety in $\optE$.
We choose a parameterized vehicle trajectory consisting of 2 parameters $k_1$ and $k_2$ which parameterize the longitudinal velocity, $x_4$, and the yaw, $x_3$, respectively.
The trajectory parameterization is given as (58) and (61) from \cite{liu2022refine} and is restated in Appendix F.
Tire forces are computed using the parameterized trajectory.

We restrict $k_1 \in [8, 10]$ to bound the longitudinal velocity of the vehicle and $k_2 \in [0.1, 1.0]$ such that only left turns are possible.
As per Remark \ref{rem:implicit_motions}, note that by parameterizing the vehicle's trajectory, we are able to implicitly encode tasks into our safe, maximally-informative trajectory computation. 
In this case, we encode left turning behaviour, however other behaviours may be encoded depending on the trajectory parameterization, and we later show an example of lane change behaviour.

The initial condition of the vehicle is $x(0) = [0, 0, 0, 0.1, 0, 0]$, meaning the vehicle starts in the low-speed regime.
As a uniform prior is incompatible with \cite{cooper2021bayesian}, we approximate the prior for their method as $\theta_i \sim \N(\hat{\theta}_i, 0.05 \hat{\theta}_i / 3)$.
We set a 5 second planning horizon with 20 measurements taken over this horizon for both both \cite{cooper2021bayesian} and our method.

Figure \ref{fig:rover_example} illustrates the optimally-informative trajectories computed using both methods.
The blue polytope represents the union of the forward reachable sets $\{\xi_j\}_{j\in\J}$ of the autonomous vehicle for all $k \in \K$, while the purple polytopes represent the forward reachable sets for the labelled trajectory parameters $k_j$. The optimal parameter $\kstar = [10, 1.0]$ and associated polytope are also shown.
The orange line represents the optimal trajectory computed using \cite{cooper2021bayesian}.

We compute the information gain using \eqref{eq:kldiv} by implementing the optimally-informative trajectories and simulating measurements along them.
We run each method in simulation 50 times, randomizing the system parameters in each trial.
Starting from the uniform prior, we apply the Dual Unscented Kalman Filter to compute the parameter posterior using simulated measurements along the computed trajectories.
The information gain \eqref{eq:kldiv} is computed using the posterior computed for each trial and the results are given in Table \ref{table:kl_compare}.

\setlength{\tabcolsep}{15pt}
\begin{table}[!t] 
\centering
\begin{tabular}{|c|c|}
\hline
\textbf{Method}  & \textbf{Information Gain} \\ \hline
Baseline {[}8{]} & 8.58 ± 2.41             \\ \hline
Ours (no safety) & 11.36 ± 2.32             \\ \hline
Our (safety)     & 7.82 ± 2.12             \\ \hline
\end{tabular}
\caption{Shannon information gain for the computed optimal trajectories using \cite{cooper2021bayesian} and our method with and without considering safety constraints. Our method computes more informative trajectories than the baseline approach as parameterizing the trajectory allows us to consider information gain over the entire time horizon. In contrast, the baseline approach greedily searches for the next best system input and as such the vehicle maneuvers end up being less informative over the time horizon. When safety is considered our approach computes less informative trajectories which is to be expected. Values shown are the means computed using \eqref{eq:theta_posterior_info_gain} over 50 trials with randomized system parameters, plus or minus one standard deviation. 
}
\label{table:kl_compare}
\end{table}

From Table \ref{table:kl_compare} we see our method computes trajectories with higher information gain than the baseline \cite{cooper2021bayesian}, even when accounting for system safety.
We believe this is because our method considers information gain over an entire parameterized trajectory whereas the baseline takes a greedy approach and only optimizes the input at each subsequent time step.
As such, we noticed that when starting from the low-speed initial condition the baseline rarely brought the vehicle into the high-speed regime.
In contrast, the optimally-informative trajectory computed using our approach always brought the vehicle into the high-speed regime to maximize information gain.

\subsubsection{Safety Comparison}

We consider a scenario where safety is essential, otherwise the vehicle may collide with other vehicles or drive off the road.
For this experiment, we do not compare against \cite{cooper2021bayesian} as their method does not consider safety when computing optimally informative trajectories.
We use the same uniform parameter prior as the last experiment.

As shown in Figure \ref{fig:main_figure}, our method is able to compute the optimally-informative, safe trajectory such that we always stay within the green polytope and remain collision free.
Like the previous example, we simulate 50 trials with randomized system parameters $\theta \in \Theta$ and present the information gain computed using \eqref{eq:kldiv} in Table \ref{table:kl_compare}.
The information gain is lower than the previous example as the safety constraint necessitates that a trajectory with lower expected information gain be chosen to avoid collisions.

\begin{figure} [!t]
    \centering
    \includegraphics[width=\columnwidth]{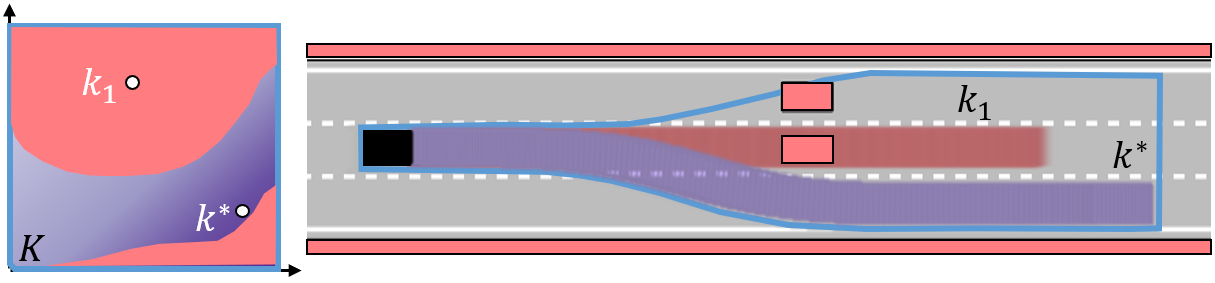}
    \caption{We demonstrate our method on the autonomous vehicle dynamics \eqref{eq:dynamics_hi} using another trajectory parameterization. Here, the vehicle performs a speed-change or lane-change maneuver, depending on the chosen $k \in \K$. The safe, optimally-informative trajectory results in the vehicle speeding up and performing a lane-change to avoid colliding with the vehicle in front of it.}
    \label{fig:lane_change_example}
\end{figure}

We also present a lane-change scenario, shown in Figure \ref{fig:lane_change_example}, using the same prior on $\theta$.
The parameterized trajectory is computed using (58) and (62) of \cite{liu2022refine} and is restated in Appendix F.
This example demonstrates that through different trajectory parameterizations we can encode different behaviours into our agent such as lane-changing or left turning.

To ensure that the computed optimally-informative trajectory maximizes the expected-information gain metric, we sample the parameter space and evaluate the metric for each sample.
The values for these samples are shown in the color map of Figures \ref{fig:main_figure}, \ref{fig:rover_example}, and \ref{fig:lane_change_example} where a darker color corresponds to a higher expected information gain value.
Thus, we conclude the parameters computed using $\optE$ are indeed generating the optimally-informative trajectories.

\section{Conclusions} \label{sec:conclusion}
We present a method for safe, differentiable Bayesian Optimal Experimental Design for dynamical systems.
Our method is able to compute optimally-informative system trajectories while maintaining system safety under mild assumptions.
Compared with state-of-the-art Bayesian Optimal Experimental Design algorithms, our method is able to compute trajectories with higher expected information gain and can do so using orders of magnitude fewer samples.

We demonstrated our method using high-fidelity vehicle dynamics and future work will include demonstrating the utility of our approach using hardware demonstrations.
Additionally, we will apply our approach to other platforms which leveraged parameterized trajectories, including drones and manipulators.
Lastly, a possible extension to this work is relaxing Assumption \ref{assum:bounded} such that we can incorporate system parameter uncertainty into the forward-reachable set computation and convert the safety constraint into a probabilistic chance constraint.

\bibliographystyle{IEEEtran}
\bibliography{references}


\end{document}